# Artificial Intelligence (AI) Based Prediction of Mortality, ICU Admission and Ventilation Support Requirement for COVID-19 Patients Using 122 Parameters

Mahbubunnabi Tamal[a,*], Mohammad Marufur Rahman[b], Maryam Alhasim[c], Mobarak Al Mulhim[d] and Mohamed Deriche[e]

[a]Department of Biomedical Engineering, College of Engineering, Imam Abdulrahman Bin Faisal University, Dammam 31441, Saudi Arabia

[b]Department of Computer Science and Engineering, Ahsanullah University of Science and Technology, Dhaka, Bangladesh

[c]Department of Radiology, College of medicine, Imam Abdulrahman bin Faisal University, Dammam 34212, Saudi Arabia

[d]Department of Critical Care, Health Cluster, Saudi Arabia

[e]Artificial Intelligence Research Centre, AIRC, Ajman University, UAE

**Corresponding Author:**
[*]Mahbubunnabi Tamal
Department of Biomedical Engineering, College of Engineering, Imam Abdulrahman Bin Faisal University, PO Box 1982, Dammam 31441, Saudi Arabia. E-mail address: mtamal@yahoo.com , mtamal@iau.edu.sa , Phone: +966 (0) 54 286 7869.

# Artificial Intelligence (AI) Based Prediction of Mortality, ICU Admission and Ventilation Support Requirement for COVID-19 Patients Using 122 Parameters

**ABSTRACT:**

For severely affected COVID-19 patients, it is crucial to identify high-risk patients and predict survival and need for intensive care (ICU). Most of the proposed models are not well reported making them less reproducible and prone to high risk of bias particularly in presence of imbalance data/class. In this study, the performances of nine machine and deep learning algorithms in combination with two widely used feature selection methods were investigated to predict "last status" representing mortality, "ICU requirement", and "ventilation days". Fivefold cross-validation was used for training and validation purposes. To minimize bias, the training and testing sets were split maintaining similar distributions. Only 10 out of 122 features were found to be useful in prediction modelling with "Acute kidney injury during hospitalization" feature being the most important one. The algorithms' performances depend on feature numbers and data pre-processing techniques. LSTM performs the best in predicting "last status" and "ICU requirement" with 90%, 92%, 86% and 95% accuracy, sensitivity, specificity, and AUC respectively. DNN performs the best in predicting "Ventilation days" with 88% accuracy. Considering all the factors and limitations including absence of exact time point of clinical onset, LSTM with carefully selected features can accurately predict "last status" and "ICU requirement". DNN performs the best in predicting "Ventilation days". Appropriate machine learning algorithm with carefully selected features and balance data can accurately predict mortality, ICU requirement and ventilation support. Such model can be very useful in emergency and pandemic where prompt and precise decision making is crucial.

**Introduction:**

The COVID-19 pandemic caused a severe global health crisis and socio-economic challenges with millions of confirmed cases and hundreds of thousands of deaths. The virus can lead to severe respiratory problems and can cause pneumonia. The severity of the disease varies from asymptomatic to severe respiratory illness requiring hospitalization and intensive care [1]. Different biomarkers have been reported to be associated with different stages of the disease [2]. The pandemic resulted in an overwhelming burden on healthcare systems worldwide, making it imperative to identify high-risk patients and predict the need for intensive care and survival. Accurate prediction of ICU admission rates and survival can play an important role in case of relapse or reoccurrence of another pandemic.

**ICU prediction:**
Several predictive models have been developed to identify patients at high risk of ICU admission. A clinical prediction model was developed incorporating age, sex, comorbidities, respiratory rate, oxygen saturation, and C-reactive protein levels to predict the need for ICU admission [3]. The study reported that the model had good predictive performance. Other studies found that age, comorbidities, and abnormal laboratory findings were significant factors associated with ICU admission rates in COVID-19 patients [4], [5]. Different machine learning models using demographic and clinical data were investigated to predict ICU admission [6], [7]. These models had sensitivity and specificity above 90% and 85% respectively indicating its high accuracy in predicting ICU admission.

**Survival prediction:**
Several studies have been conducted to identify factors associated with COVID-19 survival. A study found that age, comorbidities, and severity of illness were significant predictors of mortality in COVID-19 patients [8]. Similarly, another study reported that age, comorbidities, and the need for mechanical ventilation were significant predictors of mortality in COVID-19 patients [9]. It was reported that advanced age, high Sequential Organ Failure Assessment (SOFA) score, and D-dimer levels were associated with increased mortality in COVID-19 patients [10]. Patients with comorbidities such as hypertension, diabetes, and cardiovascular disease had a higher risk of mortality [11]. Additionally, the study found that the use of corticosteroids was associated with increased mortality in COVID-19 patients. Early diagnosis, timely treatment, and adequate supportive care were crucial for improving survival rates in COVID-19 patients [12], [13].

Advanced age, male gender, and underlying comorbidities were predictors of both ICU admission and mortality in COVID-19 patients [14]. Early recognition and prompt treatment of severe cases were crucial for improving both ICU admission rates and survival [15].

It has been reported in a review that many COVID-19 predictive models are not well reported and prone to high risk of bias [3]. In this study, the performances of seven classical machine and two deep leaning models in combination with two widely used feature selection methods (random forest and extra tree classifier) were investigated to predict "last status" representing mortality, "ICU requirement", and "ventilation days". Fivefold cross-validation was used for training and validation purposes. In each fold, 80% data were used for training the models and the rest 20% were preserved for validation. To minimize bias, the training and testing sets were split maintaining similar distributions.

**Materials and Methods:**

**Data Description:**
The publicly available data was acquired at Stony Brook University from patients who tested positive for COVID-19 [16], [17]. The dataset used in this study contains 122 demographic and clinical features of 1384 patients. The detail of all the 122 features is available in the database. Among all the 122 features, 86 are categorical and the rest 36 features are continuous numeric values. The dataset contains 3 outcome attributes – a) "last status", b) "ICU needed", and c) "ventilated days".

**Data Preprocessing and Preparation:**
Before training different machine learning algorithms, we needed to address two issues related to the data – a) Missing sample data and 2) Imbalanced data.

   a) Missing Data

Only two features (age and acute kidney injury during hospitalization) have all the samples for all patients. The rest 120 features have different numbers of sample data missing for different patients with the notation "NA (not available)". Since removal of missing samples will drastically reduce data points, they were replaced in two different ways. First, missing values of Boolean features were replaced with a fixed value (0.5). For example, sample values of the feature "pulseOx.under90" are Boolean, i.e., either true (1) and false (0). Missing sample values of this feature were replaced with 0.5 which means "do not count" by the machine learning algorithm.

Missing data of features with continuous values were handled by KNN imputation [18]. KNN imputation is a widely used effective technique for medical data imputation. The hyper-parameters of the KNN algorithm were set to the number of neighbours 5, weights "uniform", and metric "nan_euclidian".

   b) Imbalanced data

Like most of the typical medical data, the dataset used in this study is highly imbalanced. For example, only 260 samples out of 1384 samples represent death cases representing only about 19% of the total sample. To investigate the impact of imbalance in the dataset on machine learning algorithm, we experimented with both original dataset as well as normalized dataset. In the normalized dataset, samples were either generated or deleted using oversampling or undersampling techniques respectively. In oversampling, the number of samples of the minority class was increased by randomly duplicating samples using bootstrapping method so that the new oversampled dataset remained bias free. Total number of duplications made the minority class approximately half of the majority class. On the other hand, undersampling of the majority class produced a dataset that contained an almost equal number of samples of both classes by deleting samples from the majority class.

Predicting number of days for invasive ventilation ("ventilation days") support required for a patient is a regression task. Because of the availability of sparse samples with ventilation days in the dataset, all the renowned regression models performed poorly despite applying oversampling on the dataset. Because of this reason, the regression task was transformed into a classification task by binning the days into weeks. Five different bins with index 3, 4, 5, 6, 7 were produced from the original dataset. In Table 1, bin labels for these derived datasets are given and Table 2 contains the meaning of these labels. The dataset with 3 bins consists of three categorical values which are 0 (no ventilation needed), 1 (ventilation needed for about 1 week), and 2 (ventilation needed for more than 1 week). Other datasets were produced with this procedure.

**Table 1:** Bin labels of the derived datasets for ventilated days prediction

| No. of Bins | Bin Labels | | | | | | |
|---|---|---|---|---|---|---|---|
| 3 | 0 | 1 | >1 | | | | |
| 4 | 0 | 1 | 2 | >2 | | | |
| 5 | 0 | 1 | 2 | 3 | >3 | | |
| 6 | 0 | 1 | 2 | 3 | 4 | >4 | |
| 7 | 0 | 1 | 2 | 3 | 4 | 5 | >5 |

**Table 2:** Interpretation of the bin labels of Table 1 (derived datasets for ventilated days prediction)

| Symbol | Meaning |
|---|---|
| 0 | No ventilation required |
| 1 | Approximately 1 week i.e., 1 to 7 days of ventilation required |
| >1 | More than 1 week of ventilation is required |
| 2 | Approximately 2 weeks i.e., 8 to 14 days of ventilation required |
| >2 | More than 2 week of ventilation is required |
| 3 | Approximately 3 weeks i.e., 15 to 21 days of ventilation required |
| >3 | More than 3 week of ventilation is required |
| 4 | Approximately 4 weeks i.e., 22 to 28 days of ventilation required |
| >4 | More than 4 week of ventilation is required |
| 5 | Approximately 5 weeks i.e., 29 to 35 days of ventilation required |
| >5 | More than 5 week of ventilation is required |

**Automatic Feature Selection and Classification**

All the 122 features of the dataset are not equally important for predicting the outcome. Some features contribute more to the prediction than others. On the other hand, overloading the machine learning algorithm with features that are less important could impact the performance of the algorithm. Because of this reason, the most relevant features were first selected before training the algorithm. There are several techniques for feature selection. In this study, we investigated with two widely used techniques – a) random forest and b) extra tree classifier. While training and testing, we observed that predictive model trained with the features selected by the extra tree classifier performed slightly better compared to the features selected by the random forest. Though the difference is non-significant, we decided to use extra tree classifier for further investigation.

**Training and Validation:**

Top 10 features from 122 features were selected for predicting "last status", "ICU requirement", and "ventilation days". It was observed that selecting more than 10 features did not improve the prediction rather lowered the performance.

Prediction of "last status" and "ICU requirement" are binary classification tasks. For these classification tasks, seven machine learning algorithms were explored in this study. They are – 1) Random Forest (RF), 2) Logistic Regression (LR), 3) Support Vector Machine (SVM), 4) k-Nearest Neighbor (KNN), 5) XGBoost, 6) Linear Discriminant Analysis (LDA) and 7) Gaussian Naïve Bayes (NB) were used. Scikit-learn library was used to implement these algorithms. Hyper parameters used for the first six machine learning algorithms were chosen by randomized parameter search technique which are shown in Table 3.

**Table 3.** Hyper parameters used for different machine learning algorithms.

| Model | Parameters |
|---|---|
| **RF** | n_estimators = 50, criterion = entropy, max_depth = 10, min_samples_split = 8, min_samples_leaf = 1 |
| **LR** | penalty = 'l2', solver = 'newton-cg', max_iter = 20, |
| **SVM** | kernel = 'rbf', degree = 3, gamma = 'auto' |
| **KNN** | n_neighbors = 7, weights = 'distance', algorithm = 'auto', leaf_size = 10, metric = 'minkowski' |
| **XGBoost** | n_estimators = 100, learning_rate = 1.0, max_depth = 2, min_samples_split=5, min_samples_leaf=4, loss='deviance' |
| **LDA** | n_components = None, solver = 'svd' |

"RandomizedSearchCV" from the scikit-learn library was used for getting a decent parameter combination. The "n_iter "parameter of this function was set to 1000 and other parameters were kept as their default values during training.

Two deep learning approaches – 1) Deep Neural Network (DNN), and 2) Long Short-Term Memory (LSTM) were also used in this study. Figures 1(a) and 1(b) depict the LSTM and DNN architectures for "last status" and "ICU requirement" respectively used in this study. For predicting "Ventilation days", only DNN was used. The DNN architecture that was used in the study shown in Figure 1(c). The original dataset contains very few patients who required ventilation. To train the algorithm, a good amount of data is required. Because of that, only oversampled data were used for predicting "Ventilation days".

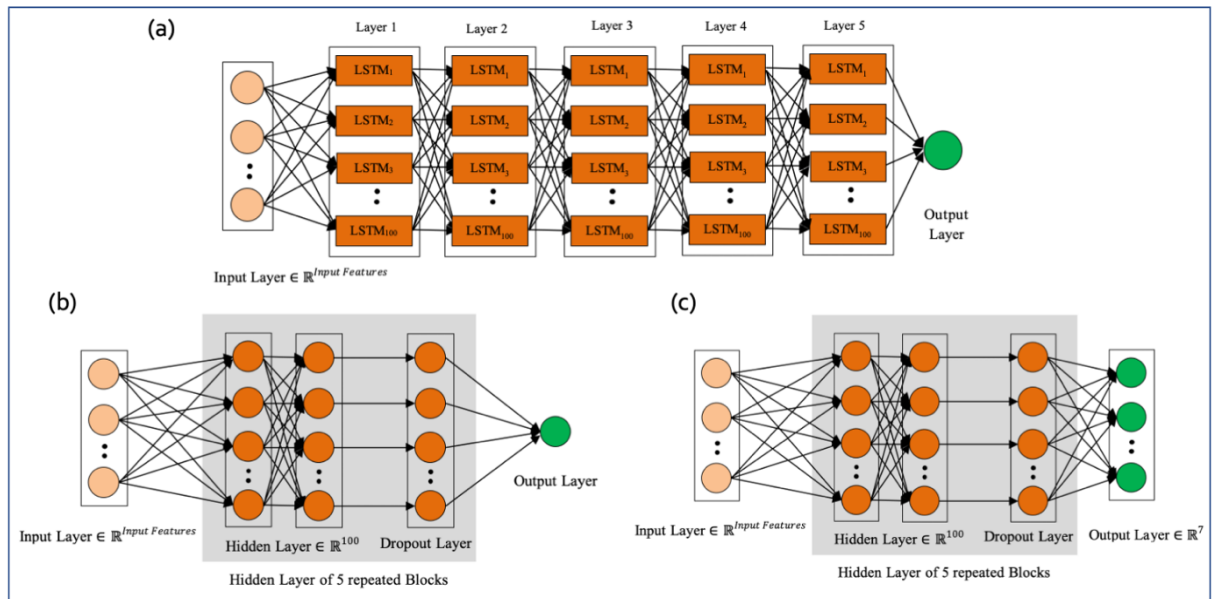

**Figure 1:** Model architecture (a) Long Short-Term Memory (LSTM) and (b) Deep Neural Network (DNN) architecture for predicting "last status'" and "ICU requirement "ICU requirement". (c) DNN architecture for predicting "Ventilation days".

Fivefold cross-validation was used for training and validation purposes. In each fold, 80% samples were used for training the models and the rest 20% of the samples were preserved for validation. Mean probability distributions of fifteen features which are present in the top ten features for the three prediction tasks are shown in Supplementary Figure 1. It is clear from the figure that the training and testing data are from the same population since they have almost the same distribution for all the features. This figure contains the probability distribution of training and testing sets of original, over-sampled and under-sampled datasets for 15 features.

The performances of the models were evaluated by measuring accuracy, sensitivity, and specificity using the following formulas:

$$\text{Accuracy} = \frac{(TP+TN)}{(TP+TN+FP+FN)} \quad (1)$$

$$\text{Sensitivity} = \frac{TP}{(TP+FN)} \quad (2)$$

$$\text{Specificity} = \frac{TN}{(TN+FP)} \quad (3)$$

TP, TN, FP, and FN stand for true positive, true negative, false positive and false negative respectively. The sensitivity of a model determines its capability of successfully predicting survival of patients among all the patients who survived and similarly specificity implies the capability of successfully predicting the death of infected patients from all the patients who passed away.

Receiver operating characteristic (ROC) curves were also generated and Area Under the Curve (AUC) were measured for each model. ROC is a graphical representation of a binary classifier's diagnostic ability. Higher the AUC of a model the better its classification efficiency.

**Results and Discussions**
**Feature selection:**
Figure 2 shows the top 10 contributing features along with their importance measure in all three prediction tasks. "Acute kidney injury during hospitalization" is the most important feature for all of them. Only four features are present in top 10 features for all three prediction models– "Blood pH below 7.5", "Gender", "Therapeutic heparin" and "Acute kidney injury during hospitalization". Gender and age are found to be vital in predicting last status. However, "Gender" has a very little impact in predicting the need of ICU, and "Ventilated days". For both these prediction cases, "Age" is not listed in the top 10 features.

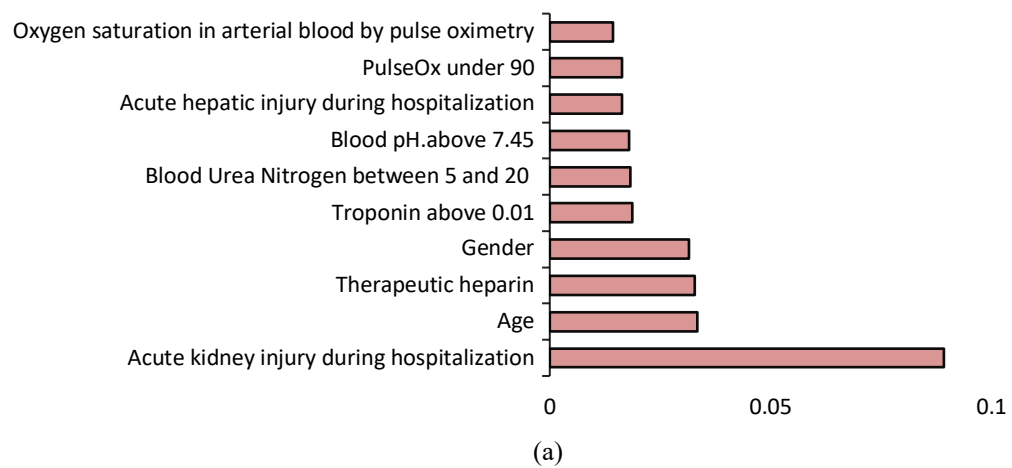

(a)

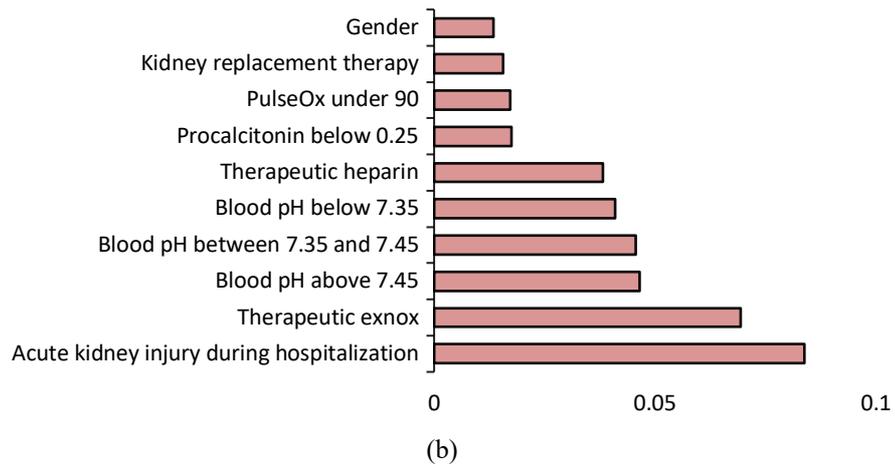

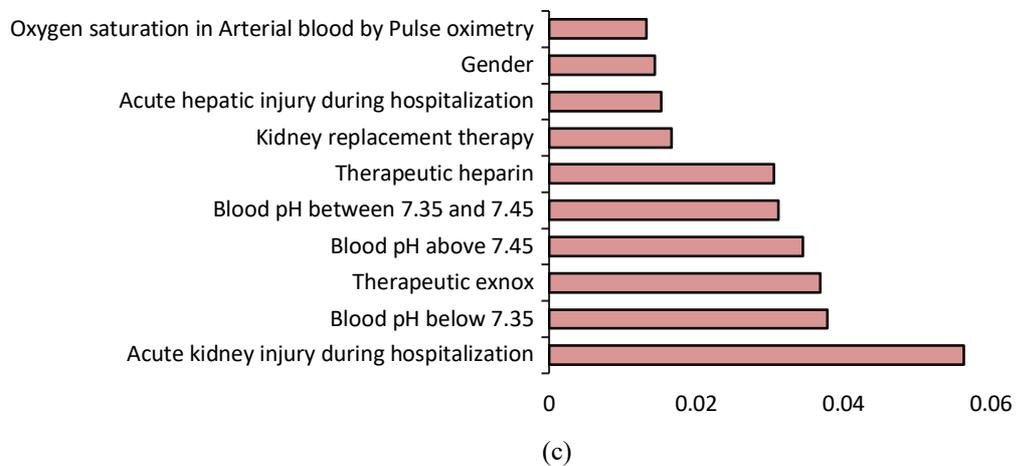

**Figure 2:** Top 10 contributing features for predicting (a) last status (b) ICU needed (c) ventilated days.

**Prediction of "last status"**

Table 4 shows the results of the top performing three models using a varied number of top-ranked features. Receiver operating characteristic (ROC) curve and Area Under the Curve (AUC) for these three models are shown in Figure 3. For the original dataset, LDA showed 90.10% (+/- 0.71%) accuracy, 93.34% (+/- 1.02%) sensitivity, and 68.83% (+/- 6.65%) specificity when top 3 features were used. Since the original dataset contains more survival cases than death cases, the specificity is low with high variance. Though XGBoost showed 91.47% (+/- 1.22%) using the top 3 features which is slightly higher than LSTM, its specificity is quite low compared to LSTM which is 63.35% (+/- 7.19%). It showed 95.75% (+/- 1.35%) sensitivity that implies models' expertise in accurately predicting survival cases. For the original dataset, in terms of accuracy SVM showed better performance than all other models. Its overall accuracy was 92.12% (+/- 0.70%) and specificity was 63.35% (+/- 6.07%) when the top 4 features were used. The highest AUC of SVM, XGBoost, LDA were 0.93, 0.95, 0.93 respectively was achieved with the top 10 features. For SVM, AUC is very much dependent on the number of features selected ranging from 0.79 to 0.93 for the top 3 and10 features respectively.

**Table 4.** Performance of top three machine learning models in 'last status' prediction using different top-ranked features from the original dataset.

| Model | Matrices | Top 3 Features | Top 4 Features | Top 5 Features | Top 6 Features | Top 8 Features | Top 10 Features |
|---|---|---|---|---|---|---|---|
| SVM | Accuracy | 91.40% (+/- 0.80%) | **92.12% (+/- 0.70%)** | 90.90% (+/- 0.61%) | 90.90% (+/- 0.62%) | 91.19% (+/- 0.67%) | 91.11% (+/- 0.66%) |
| | Sensitivity | 96.75% (+/- 1.57%) | **96.50% (+/- 1.04%)** | 95.84% (+/- 1.90%) | 96.50% (+/- 1.25%) | 97.08% (+/- 1.37%) | 96.59% (+/- 0.97%) |
| | Specificity | 56.22% (+/- 6.44%) | **63.35% (+/- 6.07%)** | 58.41% (+/- 8.93%) | 54.02% (+/- 7.04%) | 52.42% (+/- 5.41%) | 55.15% (+/- 8.71%) |
| XGBoost | Accuracy | **91.47% (+/- 1.22%)** | 91.47% (+/- 1.78%) | 91.47% (+/- 1.32%) | 91.19% (+/- 1.45%) | 91.33% (+/- 1.68%) | 91.62% (+/- 1.28%) |
| | Sensitivity | **95.75% (+/- 1.35%)** | 96.17% (+/- 1.27%) | 96.09% (+/- 1.33%) | 95.92% (+/- 1.30%) | 96.42% (+/- 1.07%) | 96.17% (+/- 1.22%) |
| | Specificity | **63.35% (+/- 7.19%)** | 60.62% (+/- 10.49%) | 61.16% (+/- 8.49%) | 60.08% (+/- 9.44%) | 57.90% (+/- 12.12%) | 61.70% (+/- 6.02%) |
| LDA | Accuracy | **90.10% (+/- 0.71%)** | 89.45% (+/- 0.74%) | 89.31% (+/- 0.78%) | 89.31% (+/- 0.84%) | 90.03% (+/- 0.66%) | 89.89% (+/- 1.22%) |
| | Sensitivity | **93.34% (+/- 1.02%)** | 93.01% (+/- 1.54%) | 93.17% (+/- 1.67%) | 93.09% (+/- 1.77%) | 93.76% (+/- 1.80%) | 93.51% (+/- 1.75%) |
| | Specificity | **68.83% (+/- 6.65%)** | 66.13% (+/- 4.67%) | 63.95% (+/- 5.95%) | 64.50% (+/- 5.79%) | 65.59% (+/- 9.10%) | 66.16% (+/- 5.09%) |

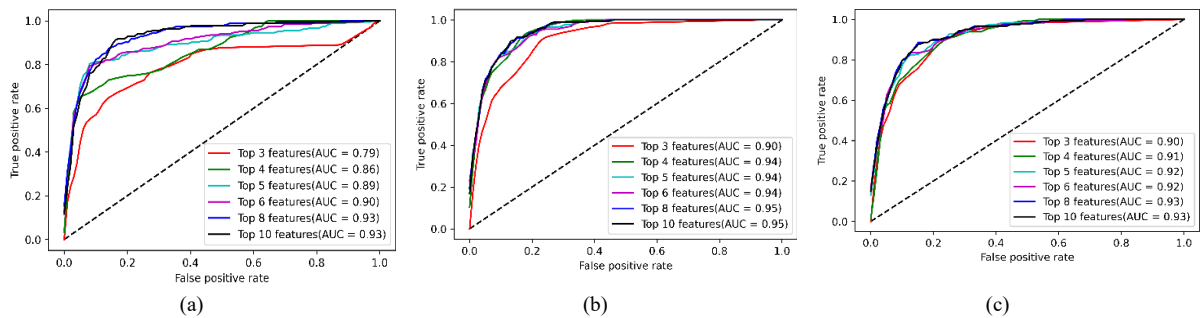

**Figure 3:** ROC curve and AUC of top three performing models (a) SVM (b) XGBoost (c) LDA

For under-sampled and over-sampled dataset, the specificity increases significantly as the data between survival and death becomes balance though the accuracy and sensitivity decrease compared to the original data. RF, XGBoost and LSTM display the best performance compared to other models using top 5 features for under-sampled data. The performance of these three methods with the 5 features are comparable as shown in Table 5 and Figure 4 (near to 90% accuracy, sensitivity, specificity and

**Table 5.** Performance of top three machine learning models in 'last status' prediction using top-ranked features from the under-sampled dataset.

| Model | Matrices | Top 3 Features | Top 4 Features | Top 5 Features | Top 6 Features | Top 8 Features | Top 10 Features |
|---|---|---|---|---|---|---|---|
| RF | Accuracy | 78.69% (+/- 2.01%) | 86.40% (+/- 2.24%) | **87.58% (+/- 2.23%)** | 85.71% (+/- 4.17%) | 84.55% (+/- 4.00%) | 84.30% (+/- 5.80%) |
| | Sensitivity | 85.67% (+/- 5.13%) | 86.90% (+/- 2.71%) | **86.51% (+/- 6.11%)** | 85.30% (+/- 8.97%) | 86.86% (+/- 5.02%) | 84.40% (+/- 5.38%) |
| | Specificity | 69.38% (+/- 7.58%) | 85.78% (+/- 6.54%) | **89.10% (+/- 7.43%)** | 86.38% (+/- 9.22%) | 81.49% (+/- 5.08%) | 84.17% (+/- 7.13%) |
| XGBoost | Accuracy | 79.15% (+/- 1.40%) | 86.41% (+/- 3.06%) | **88.76% (+/- 1.78%)** | 87.11% (+/- 4.90%) | 86.17% (+/- 4.87%) | 88.27% (+/- 6.25%) |
| | Sensitivity | 84.45% (+/- 4.35%) | 88.54% (+/- 2.73%) | **89.38% (+/- 4.53%)** | 90.19% (+/- 7.22%) | 85.63% (+/- 4.59%) | 89.73% (+/- 5.08%) |
| | Specificity | 72.15% (+/- 7.91%) | 83.60% (+/- 8.11%) | **87.99% (+/- 4.70%)** | 83.11% (+/- 8.94%) | 86.91% (+/- 5.99%) | 86.38% (+/- 8.52%) |
| LSTM | Accuracy | 79.39% (+/- 3.54%) | 85.49% (+/- 2.76%) | **89.93% (+/- 2.85%)** | 87.59% (+/- 3.01%) | 87.35% (+/- 4.57%) | 88.76% (+/- 2.40%) |
| | Sensitivity | 87.30% (+/- 3.93%) | 88.10% (+/- 7.28%) | **89.37% (+/- 5.36%)** | 85.65% (+/- 2.58%) | 85.62% (+/- 6.32%) | 92.62% (+/- 4.94%) |
| | Specificity | 68.87% (+/- 9.40%) | 81.92% (+/- 8.68%) | **90.71% (+/- 2.84%)** | 90.14% (+/- 6.59%) | 89.61% (+/- 3.20%) | 83.62% (+/- 5.92%) |

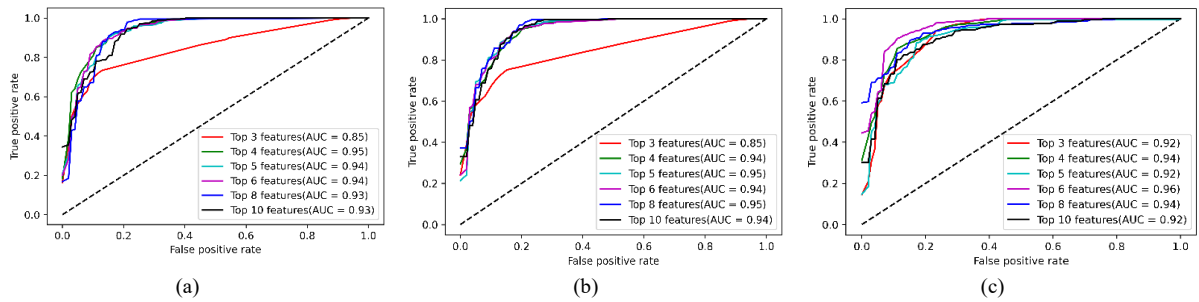

**Figure 4:** ROC curve and AUC of top three performing models (a) RF (b) XGBoost (c) LSTM

For oversampled data, RF, DNN and LSTM performs the best and their performances are comparable with more than 90% accuracy, sensitivity, specificity, and AUC as shown in Table 6 and Figure 5. This accuracy is achieved with top 8 features for RF and top 10 features for DNN and LSTM.

**Table 6.** Performance of top three machine learning models in 'last status' prediction using top-ranked features from the oversampled dataset.

| Model | Matrices | Top 3 Features | Top 4 Features | Top 5 Features | Top 6 Features | Top 8 Features | Top 10 Features |
|---|---|---|---|---|---|---|---|
| RF | Accuracy | 82.95% (+/- 0.91%) | 86.12% (+/- 2.07%) | 89.89% (+/- 1.10%) | 89.95% (+/- 1.63%) | **92.00% (+/- 1.07%)** | 91.84% (+/- 0.60%) |
| | Sensitivity | 89.43% (+/- 1.91%) | 89.26% (+/- 4.30%) | 90.84% (+/- 2.02%) | 91.51% (+/- 1.76%) | **91.76% (+/- 1.38%)** | 91.51% (+/- 1.48%) |
| | Specificity | 70.00% (+/- 3.37%) | 79.83% (+/- 6.02%) | 88.00% (+/- 6.09%) | 86.83% (+/- 3.05%) | **92.50% (+/- 4.28%)** | 92.50% (+/- 3.46%) |
| DNN | Accuracy | 84.18% (+/- 1.88%) | 87.28% (+/- 0.71%) | 90.23% (+/- 1.52%) | 89.45% (+/- 1.45%) | 92.28% (+/- 1.78%) | **93.95% (+/- 1.27%)** |
| | Sensitivity | 90.34% (+/- 3.37%) | 94.42% (+/- 1.43%) | 92.84% (+/- 2.36%) | 93.92% (+/- 2.48%) | 94.01% (+/- 2.52%) | **95.01% (+/- 1.01%)** |
| | Specificity | 71.83% (+/- 9.24%) | 73.00% (+/- 2.56%) | 85.00% (+/- 5.53%) | 80.50% (+/- 5.69%) | 88.83% (+/- 2.56%) | **91.83% (+/- 4.64%)** |
| LSTM | Accuracy | 83.01% (+/- 0.87%) | 88.17% (+/- 1.27%) | 90.67% (+/- 1.20%) | 90.12% (+/- 1.92%) | 91.40% (+/- 2.66%) | **93.01% (+/- 2.50%)** |
| | Sensitivity | 90.09% (+/- 1.09%) | 91.01% (+/- 2.26%) | 91.17% (+/- 1.81%) | 93.01% (+/- 1.87%) | 91.43% (+/- 3.01%) | **93.17% (+/- 2.83%)** |
| | Specificity | 68.83% (+/- 2.67%) | 82.50% (+/- 6.21%) | 89.67% (+/- 3.86%) | 84.33% (+/- 4.70%) | 91.33% (+/- 3.48%) | **92.67% (+/- 5.97%)** |

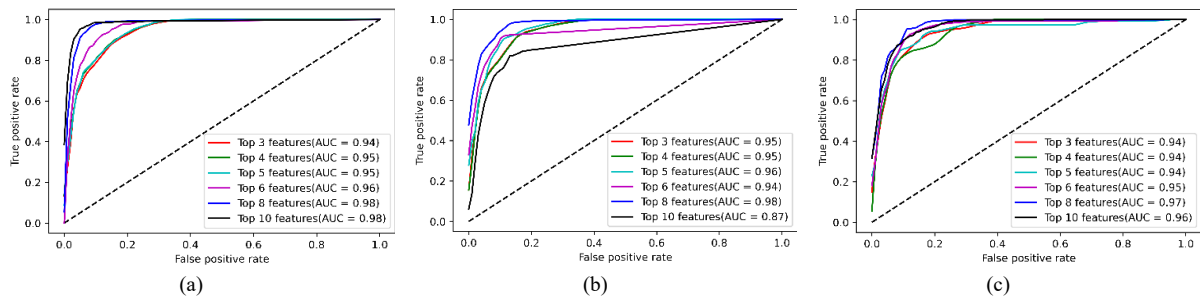

**Figure 5:** ROC curve and AUC of top three performing models (a) RF (b) DNN (c) LSTM

None of the algorithm is in top 3 positions for all original, under-sampled and oversampled data. XGBoost is common for original and under-sampled data, whereas LSTM is common for under-sampled and oversampled data.

**Prediction of "ICU Requirement"**
Among all the algorithms XGBoost, LDA and LSTM are the top three performers in predicting "ICU Requirement". Like "last status", accuracy and sensitivity for all three methods are above

90% for original data. However, specificity is around 70%. The results are shown in Table 7. ROC curves along with the AUC is shown in Figure 6. Number of top features does not impact the outcome of accuracy, sensitivity, and AUC significantly. However, as the number of features increases, specificity decreases for all the three machine learning algorithms.

**Table 7.** Performance comparison of top three machine learning models in predicting "ICU Requirement" using different number of features of the original data.

| Model | Matrices | Top 3 Features | Top 4 Features | Top 5 Features | Top 6 Features | Top 8 Features | Top 10 Features |
|---|---|---|---|---|---|---|---|
| XGBoost | Accuracy | 90.03% (+/- 0.87%) | 90.03% (+/- 0.87%) | **90.25% (+/- 0.94%)** | 90.25% (+/- 0.94%) | 89.38% (+/- 1.30%) | 88.73% (+/- 1.28%) |
| | Sensitivity | 95.11% (+/- 1.49%) | 95.11% (+/- 1.49%) | **95.11% (+/- 1.49%)** | 95.11% (+/- 1.49%) | 95.37% (+/- 1.66%) | 94.84% (+/- 1.75%) |
| | Specificity | 68.08% (+/- 8.12%) | 68.08% (+/- 8.12%) | **69.23% (+/- 8.85%)** | 69.23% (+/- 8.85%) | 63.46% (+/- 11.54%) | 62.31% (+/- 9.78%) |
| LDA | Accuracy | 84.32% (+/- 1.73%) | 87.57% (+/- 0.73%) | 88.08% (+/- 1.81%) | **90.03% (+/- 0.81%)** | 89.23% (+/- 0.99%) | 89.60% (+/- 1.15%) |
| | Sensitivity | 91.10% (+/- 1.73%) | 96.26% (+/- 1.00%) | 95.73% (+/- 1.15%) | **94.66% (+/- 1.26%)** | 94.48% (+/- 1.56%) | 94.75% (+/- 1.75%) |
| | Specificity | 55.00% (+/- 5.78%) | 50.00% (+/- 5.70%) | 55.00% (+/- 8.90%) | **70.00% (+/- 8.82%)** | 66.54% (+/- 9.47%) | 67.31% (+/- 9.58%) |
| LSTM | Accuracy | 87.64% (+/- 1.44%) | 89.60% (+/- 1.38%) | 89.81% (+/- 1.31%) | **90.17% (+/- 0.89%)** | 90.03% (+/- 1.30%) | 90.10% (+/- 1.37%) |
| | Sensitivity | 95.91% (+/- 1.47%) | 94.04% (+/- 1.46%) | 95.55% (+/- 1.35%) | **94.93% (+/- 1.37%)** | 95.28% (+/- 1.37%) | 95.46% (+/- 1.61%) |
| | Specificity | 51.92% (+/- 2.43%) | 70.38% (+/- 9.85%) | 65.00% (+/- 8.01%) | **69.62% (+/- 8.80%)** | 67.31% (+/- 9.18%) | 66.92% (+/- 5.88%) |

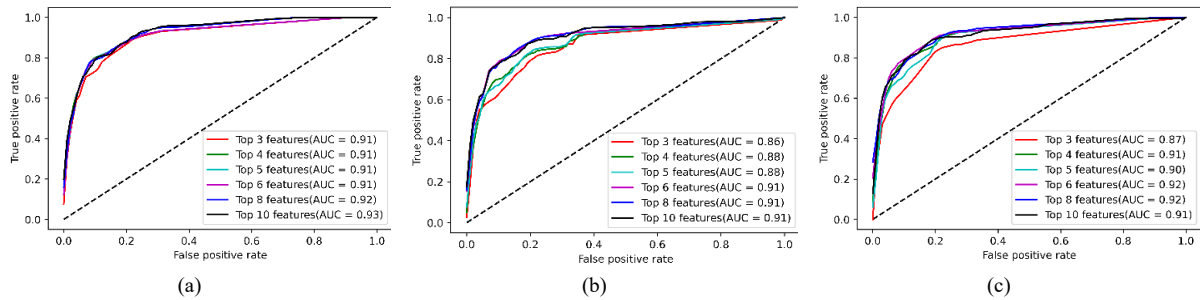

**Figure 6:** ROC curve and AUC of top three performing models for predicting "ICU Requirement" using original data. (a) XGBoost, (b) LDA and (c) LSTM

For the under-sampled data, SVM, DNN and LSTM are the top three performer using 10 features (Table 8). DNN provides the highest AUC of 95% with top 10 features, followed by LSTM with 93% AUC and top 10 features (Figure 7).

**Table 8.** Performance comparison of top three machine learning models in predicting "ICU Requirement" using different number of features of the under-sampled data.

| Model | Matrices | Top 3 Features | Top 4 Features | Top 5 Features | Top 6 Features | Top 8 Features | Top 10 Features |
|---|---|---|---|---|---|---|---|
| SVM | Accuracy | 81.02% (+/- 3.02%) | 84.82% (+/- 2.89%) | 84.32% (+/- 2.44%) | 84.32% (+/- 2.44%) | 84.32% (+/- 2.44%) | **86.31% (+/- 1.21%)** |
| | Sensitivity | 78.62% (+/- 4.39%) | 84.11% (+/- 4.55%) | 84.11% (+/- 4.55%) | 84.11% (+/- 4.55%) | 84.40% (+/- 4.10%) | **86.41% (+/- 4.27%)** |
| | Specificity | 84.23% (+/- 10.97%) | 85.77% (+/- 8.48%) | 84.62% (+/- 7.98%) | 84.62% (+/- 7.98%) | 84.23% (+/- 7.24%) | **86.15% (+/- 5.22%)** |
| DNN | Accuracy | 85.32% (+/- 1.87%) | 83.66% (+/- 3.79%) | 85.64% (+/- 2.13%) | 86.31% (+/- 2.46%) | 85.81% (+/- 1.92%) | **89.28% (+/- 2.71%)** |
| | Sensitivity | 92.53% (+/- 7.83%) | 94.22% (+/- 1.83%) | 93.64% (+/- 1.97%) | 94.22% (+/- 0.92%) | 95.38% (+/- 1.68%) | **97.98% (+/- 2.17%)** |
| | Specificity | 75.77% (+/- 8.30%) | 69.62% (+/- 6.48%) | 75.00% (+/- 4.39%) | 75.77% (+/- 5.25%) | 73.08% (+/- 4.03%) | **77.69% (+/- 3.57%)** |
| LSTM | Accuracy | 86.64% (+/- 3.23%) | 81.84% (+/- 2.49%) | 82.01% (+/- 3.22%) | 85.48% (+/- 1.99%) | 86.80% (+/- 2.16%) | **87.30% (+/- 3.45%)** |
| | Sensitivity | 85.84% (+/- 4.69%) | 78.31% (+/- 3.95%) | 76.29% (+/- 4.77%) | 84.37% (+/- 3.65%) | 87.55% (+/- 6.07%) | **89.30% (+/- 3.40%)** |
| | Specificity | 87.69% (+/- 4.14%) | 86.54% (+/- 5.57%) | 89.62% (+/- 3.12%) | 86.92% (+/- 6.25%) | 85.77% (+/- 6.39%) | **84.62% (+/- 7.98%)** |

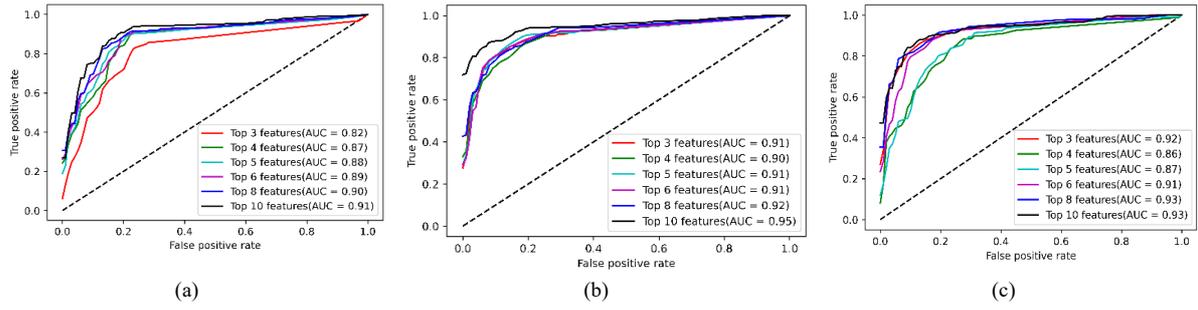

**Figure 7:** ROC curve and AUC of top three performing models for predicting "ICU Requirement" using under-sampled data. (a) SVM, (b) DNN and (c) LSTM.

Accuracy, sensitivity, and specificity are shown in Table 9 for the oversampled data. In this case of oversampled data, RF, SVM and LSTM are the top three performer using 10 features., The highest AUC of 95% is provided by the LSTM algorithm with top 10 features, followed by RF with 93% AUC and top 10 features as shown in Figure 8. LSTM is the most robust across original, under-sampled and oversampled data and perform the best in predicting "ICU Requirement".

**Table 9.** Performance comparison of top three machine learning models in predicting "ICU Requirement" using different number of features of the oversampled data.

| Model | Matrices | Top 3 Features | Top 4 Features | Top 5 Features | Top 6 Features | Top 8 Features | Top 10 Features |
|---|---|---|---|---|---|---|---|
| RF | Accuracy | 86.12% (+/- 1.45%) | 86.54% (+/- 2.07%) | 86.77% (+/- 1.82%) | 85.65% (+/- 1.63%) | 85.70% (+/- 1.42%) | **87.66% (+/- 2.05%)** |
| | Sensitivity | 93.50% (+/- 2.05%) | 93.77% (+/- 1.85%) | 92.97% (+/- 2.12%) | 91.90% (+/- 1.73%) | 91.99% (+/- 1.65%) | **92.08% (+/- 2.10%)** |
| | Specificity | 71.36% (+/- 1.46%) | 72.06% (+/- 3.12%) | 74.38% (+/- 2.30%) | 73.14% (+/- 3.06%) | 73.12% (+/- 2.44%) | **78.82% (+/- 3.21%)** |
| SVM | Accuracy | 86.12% (+/- 1.45%) | 86.54% (+/- 2.07%) | 87.01% (+/- 1.54%) | 85.53% (+/- 1.64%) | 85.82% (+/- 1.95%) | **86.65% (+/- 1.54%)** |
| | Sensitivity | 93.50% (+/- 2.05%) | 93.77% (+/- 1.85%) | 93.42% (+/- 1.68%) | 91.81% (+/- 2.05%) | 90.48% (+/- 2.60%) | **90.92% (+/- 2.22%)** |
| | Specificity | 71.36% (+/- 1.46%) | 72.06% (+/- 3.12%) | 74.20% (+/- 2.15%) | 72.97% (+/- 3.66%) | 76.50% (+/- 3.58%) | **78.11% (+/- 1.96%)** |
| LSTM | Accuracy | 86.77% (+/- 2.03%) | 86.12% (+/- 2.21%) | 87.37% (+/- 1.30%) | 87.84% (+/- 2.09%) | 88.73% (+/- 1.49%) | **89.86% (+/- 1.32%)** |
| | Sensitivity | 94.13% (+/- 1.94%) | 94.31% (+/- 1.90%) | 94.31% (+/- 1.81%) | 93.32% (+/- 2.23%) | 94.22% (+/- 1.93%) | **94.13% (+/- 1.60%)** |
| | Specificity | 72.08% (+/- 4.20%) | 69.75% (+/- 3.31%) | 73.48% (+/- 1.68%) | 76.85% (+/- 5.24%) | 77.76% (+/- 1.57%) | **81.32% (+/- 3.44%)** |

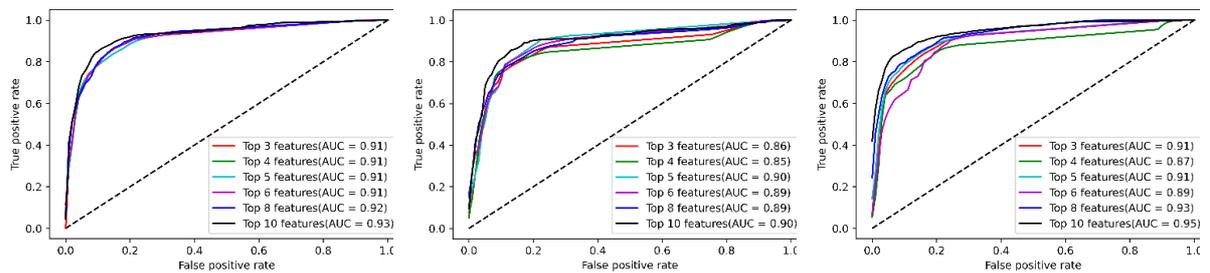

**Figure 8:** ROC curve and AUC of top three performing models for predicting "ICU Requirement" using oversampled data. (a) RF, (b) SVM and (c) LSTM.

**Prediction of "Ventilation Days"**

Performance of the DNN model for these datasets using different number of features are shown in Table 10. Including 10 features increase the prediction performance. As the number of bins increases (bins are defined in Table 1 and 2), the performance of the algorithm decreases. Among all the datasets, the dataset with 3 bins (0 week or no need for ventilation, 1 week, and more than 1 week) showed the highest accuracy of 87.59% with the top 10 feature. For 7 bins, the model achieved 75.85% accuracy. In Figure 9, confusion matrices for all the datasets and for different number of features with highest accuracy are shown.

**Table 10.** Percentage accuracy of different models in estimating 'Ventilated Days' in terms of weeks.

| No. of Classes | Top 3 Features | Top 4 Features | Top 5 Features | Top 6 Features | Top 8 Features | Top 10 Features |
|---|---|---|---|---|---|---|
| 7 | 72.27% (+/- 0.64%) | 70.56% (+/- 1.20%) | 72.10% (+/- 1.77%) | 73.01% (+/- 1.39%) | 75.29% (+/- 1.57%) | **75.85% (+/- 1.49%)** |
| 6 | 71.07% (+/- 0.92%) | 71.93% (+/- 1.09%) | 72.32% (+/- 2.06%) | 73.63% (+/- 1.21%) | **77.68% (+/- 1.55%)** | 75.57% (+/- 2.93%) |
| 5 | 73.06% (+/- 1.49%) | 71.98% (+/- 1.65%) | 76.31% (+/- 1.27%) | 74.83% (+/- 0.58%) | 76.99% (+/- 1.34%) | **82.35% (+/- 1.74%)** |
| 4 | 75.17% (+/- 1.61%) | 76.08% (+/- 2.47%) | 76.82% (+/- 1.99%) | 78.42% (+/- 1.58%) | 80.41% (+/- 1.19%) | **83.03% (+/- 1.36%)** |
| 3 | 80.92% (+/- 1.75%) | 83.89% (+/- 1.57%) | 84.40% (+/- 1.64%) | 84.11% (+/- 1.82%) | 85.65% (+/- 1.66%) | **87.59% (+/- 2.54%)** |

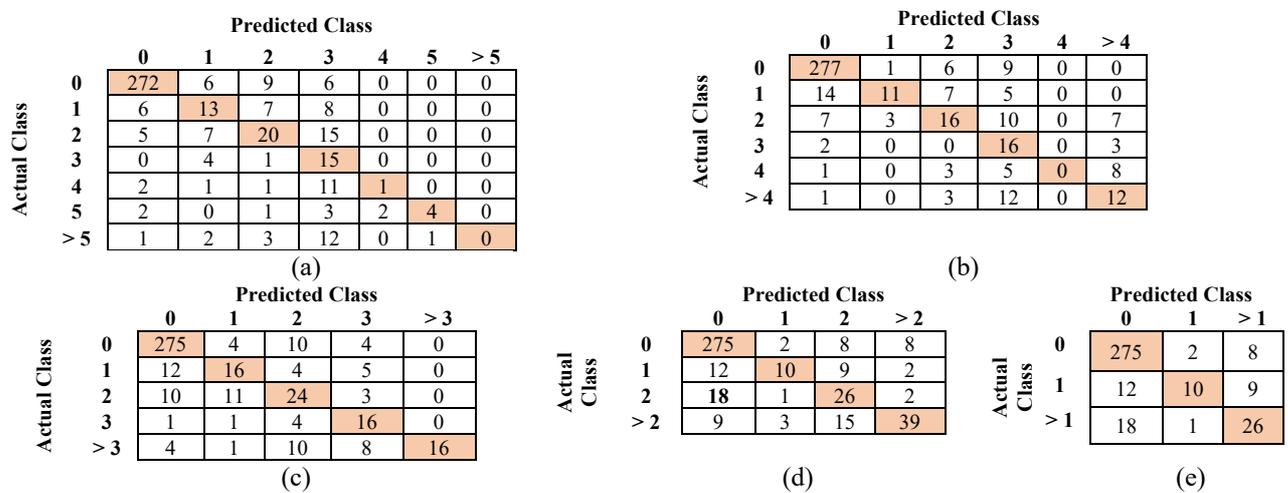

**Figure 9:** Confusion Matrix of ventilated days prediction (a) 7 class 10 features (b) 6 class 8 features (c) 5 class 10 features (d) 4 class 10 features (e) 3 class 10 features

**Discussion:**

In this study, automatic feature selection and prediction models of three outcomes 1) "last status" representing mortality, 2) "ICU requirement", and 3) "ventilation days" were investigated using different machine learning algorithms.

Predictive models for COVID-19 outcomes rely on diverse datasets encompassing a wide range of features. Feature selection plays a crucial role in machine learning especially when dealing with limited datasets, influencing model performance, interpretability, and computational efficiency. Appropriate selection of features is crucial to get an accurate prediction using machine learning algorithms [19]. Selection of too many or too less features adversely affect the performance of algorithm. In this study, it is found that the importance of most of the clinical features is related to the type of outcome predictions except "Acute kidney injury during hospitalization". This feature is strongly associated with severe forms of COVID-19. The presence of "acute kidney injury" serves as an indicator of disease severity and may contribute to risk stratification and predicting adverse outcomes, including increased mortality rates among COVID-19 patients [20]–[22].

"Blood pH" presents a decent discrimination capability especially in predicting "ICU requirement", and "ventilated days". Similar finding has been reported indicating that a decrease in blood pH might be indicative of a more severe clinical course in COVID-19 patients [23], [24]. However, the exact mechanisms by which COVID-19 affects blood pH are not fully understood, and the relationship between blood pH and disease severity is likely influenced by various factors, including respiratory function, immune response, and comorbidities. Gender and age are the other two most prominent features in predicting "last status". Similar finding has been reported by different studies [25]–[28].

Since the original dataset consists of more survival data than data of patients who died (i.e., limited representation of positive cases), the performance of almost all the models is good at predicting patients' survival and hence shows very high sensitivity. On the other hand, because of the limited number of patients who died in the training dataset the machine learning models struggle in predicting death cases. The scarcity of severe or critical outcome instances compared to mild or asymptomatic cases can lead to biased models that may not adequately capture the factors influencing severe disease progression [29]. Imbalanced datasets compromise the sensitivity and specificity of outcome prediction models, impacting their ability to correctly identify and differentiate between different outcome classes [30], [31].

Various sampling techniques such as oversampling the minority class, undersampling the majority class, or employing more advanced methods like SMOTE (Synthetic Minority Over-sampling Technique) can address class imbalance and improve the ability of the model to predict less frequent outcomes [32], [33]. Using the under-sampled dataset (i.e., ensuring class balance) in this study, capabilities of all the models in accurately predicting death cases increased. As the data increases, the performances of all the method increase. With the oversampled data, the performance of different machine learning algorithm increases with the increase of number of features. No single algorithm is stands out in performance for all three data cases implying that the performance of machine algorithm is very sensitive to the data imbalance.

Apart from imbalance distribution of the data, the other main limitation of this dataset is that the exact time point of the collection of the clinical data is not available and hence, it was difficult to set a reference time point to compare different clinical parameters of different patients. Time of admission is typically considered as reference point for prediction modelling for COVID-19 without explicitly accounting for history of the vital clinical markers [34], [35]. Saria, S. et al (2021) [36] investigated prediction modelling using clinical onset. However, they reported wide variation in time between the first alert and onset of Acute Respiratory Failure (ARF) that makes machine learning based prediction modeling more challenging.

**Conclusion:**
In conclusion, predicting ICU admission rates and survival is crucial in managing the COVID-19 pandemic. Age, comorbidities, abnormal laboratory findings, and severity of illness have been identified as significant factors associated with ICU admission rates and survival in COVID-19 patients. Overall, considering all the factors and limitations, Long Short-Term Memory (LSTM) with carefully selected features can accurately predict "last status" and "ICU requirement" within 90% accuracy, sensitivity, and specificity. On the other hand, DNN performs the best in predicting "Ventilation days". Early recognition, prompt treatment, and adequate supportive care are essential for improving ICU admission rates and survival in COVID-19 patients. These findings can guide healthcare professionals in managing COVID-19 patients and reducing the burden on healthcare systems.

Reference:
[1] B. Hu, H. Guo, P. Zhou, and Z.-L. Shi, "Characteristics of SARS-CoV-2 and COVID-19," *Nat Rev Microbiol*, vol. 19, no. 3, pp. 141–154, Mar. 2021, doi: 10.1038/s41579-020-00459-7.
[2] Y. Tjendra *et al.*, "Predicting Disease Severity and Outcome in COVID-19 Patients: A Review of Multiple Biomarkers," *Arch Pathol Lab Med*, vol. 144, no. 12, pp. 1465–1474, Dec. 2020, doi: 10.5858/arpa.2020-0471-SA.
[3] L. Wynants *et al.*, "Prediction models for diagnosis and prognosis of covid-19: systematic review and critical appraisal," *BMJ*, p. m1328, Apr. 2020, doi: 10.1136/bmj.m1328.
[4] S. Alhumaid *et al.*, "Clinical features and prognostic factors of intensive and non-intensive 1014 COVID-19 patients: an experience cohort from Alahsa, Saudi Arabia," *Eur J Med Res*, vol. 26, no. 1, p. 47, Dec. 2021, doi: 10.1186/s40001-021-00517-7.
[5] J. E. Machado-Alba *et al.*, "Factors associated with admission to the intensive care unit and mortality in patients with COVID-19, Colombia," *PLoS One*, vol. 16, no. 11, p. e0260169, Nov. 2021, doi: 10.1371/journal.pone.0260169.
[6] S. Saadatmand, K. Salimifard, R. Mohammadi, A. Kuiper, M. Marzban, and A. Farhadi, "Using machine learning in prediction of ICU admission, mortality, and length of stay in the early stage of admission of COVID-19 patients," *Ann Oper Res*, Sep. 2022, doi: 10.1007/s10479-022-04984-x.
[7] T. Dan *et al.*, "Machine Learning to Predict ICU Admission, ICU Mortality and Survivors' Length of Stay among COVID-19 Patients: Toward Optimal Allocation of ICU Resources," in *2020 IEEE International Conference on Bioinformatics and Biomedicine (BIBM)*, IEEE, Dec. 2020, pp. 555–561. doi: 10.1109/BIBM49941.2020.9313292.
[8] T. Chen *et al.*, "Clinical characteristics of 113 deceased patients with coronavirus disease 2019: Retrospective study," *The BMJ*, vol. 368, Mar. 2020, doi: 10.1136/bmj.m1091.
[9] G. Grasselli *et al.*, "Baseline Characteristics and Outcomes of 1591 Patients Infected With SARS-CoV-2 Admitted to ICUs of the Lombardy Region, Italy," *JAMA*, vol. 323, no. 16, p. 1574, Apr. 2020, doi: 10.1001/jama.2020.5394.
[10] Z. Yang, Q. Hu, F. Huang, S. Xiong, and Y. Sun, "The prognostic value of the SOFA score in patients with COVID-19," *Medicine*, vol. 100, no. 32, p. e26900, Aug. 2021, doi: 10.1097/MD.0000000000026900.
[11] C. Huang *et al.*, "Clinical features of patients infected with 2019 novel coronavirus in Wuhan, China," *The Lancet*, vol. 395, no. 10223, pp. 497–506, Feb. 2020, doi: 10.1016/S0140-6736(20)30183-5.
[12] E. Abbasi-Oshaghi, F. Mirzaei, F. Farahani, I. Khodadadi, and H. Tayebinia, "Diagnosis and treatment of coronavirus disease 2019 (COVID-19): Laboratory, PCR, and chest CT imaging findings," *International Journal of Surgery*, vol. 79, pp. 143–153, Jul. 2020, doi: 10.1016/j.ijsu.2020.05.018.
[13] Q. Sun, H. Qiu, M. Huang, and Y. Yang, "Lower mortality of COVID-19 by early recognition and intervention: experience from Jiangsu Province," *Ann Intensive Care*, vol. 10, no. 1, p. 33, Dec. 2020, doi: 10.1186/s13613-020-00650-2.
[14] S. Richardson *et al.*, "Presenting Characteristics, Comorbidities, and Outcomes Among 5700 Patients Hospitalized With COVID-19 in the New York City Area," *JAMA*, vol. 323, no. 20, p. 2052, May 2020, doi: 10.1001/jama.2020.6775.

[15] L. A. Hajjar *et al.*, "Intensive care management of patients with COVID-19: a practical approach," *Ann Intensive Care*, vol. 11, no. 1, p. 36, Dec. 2021, doi: 10.1186/s13613-021-00820-w.

[16] "Stony Brook University COVID-19 Positive Cases (COVID-19-NY-SBU) - The Cancer Imaging Archive (TCIA) Public Access - Cancer Imaging Archive Wiki." Accessed: May 01, 2023. [Online]. Available: https://wiki.cancerimagingarchive.net/pages/viewpage.action?pageId=89096912

[17] K. Clark *et al.*, "The cancer imaging archive (TCIA): Maintaining and operating a public information repository," *J Digit Imaging*, vol. 26, no. 6, pp. 1045–1057, Dec. 2013, doi: 10.1007/S10278-013-9622-7.

[18] J. R. Stevens, A. Suyundikov, and M. L. Slattery, "Accounting for Missing Data in Clinical Research," *JAMA*, vol. 315, no. 5, p. 517, Feb. 2016, doi: 10.1001/jama.2015.16461.

[19] I. Guyon and A. Elisseeff, "An Introduction to Variable and Feature Selection," *Journal of Machine Learning*, vol. 3, pp. 1157–1182, 2003.

[20] S. Gupta *et al.*, "AKI Treated with Renal Replacement Therapy in Critically Ill Patients with COVID-19," *Journal of the American Society of Nephrology*, vol. 32, no. 1, pp. 161–176, Jan. 2021, doi: 10.1681/ASN.2020060897.

[21] J. J. Ng, Y. Luo, K. Phua, and A. M. T. L. Choong, "Acute kidney injury in hospitalized patients with coronavirus disease 2019 (COVID-19): A meta-analysis," *Journal of Infection*, vol. 81, no. 4, pp. 647–679, Oct. 2020, doi: 10.1016/j.jinf.2020.05.009.

[22] A. Izcovich *et al.*, "Prognostic factors for severity and mortality in patients infected with COVID-19: A systematic review," *PLoS One*, vol. 15, no. 11, p. e0241955, Nov. 2020, doi: 10.1371/journal.pone.0241955.

[23] T. Ilczak, A. Micor, W. Waksmańska, R. Bobiński, and M. Kawecki, "Factors which impact the length of hospitalisation and death rate of COVID-19 patients based on initial triage using capillary blood gas tests: a single centre study," *Sci Rep*, vol. 12, no. 1, p. 17458, Oct. 2022, doi: 10.1038/s41598-022-22388-6.

[24] J. Hu *et al.*, "Detection of COVID-19 severity using blood gas analysis parameters and Harris hawks optimized extreme learning machine," *Comput Biol Med*, vol. 142, p. 105166, Mar. 2022, doi: 10.1016/j.compbiomed.2021.105166.

[25] C. M. Petrilli *et al.*, "Factors associated with hospital admission and critical illness among 5279 people with coronavirus disease 2019 in New York City: prospective cohort study," *BMJ*, p. m1966, May 2020, doi: 10.1136/bmj.m1966.

[26] C. Gebhard, V. Regitz-Zagrosek, H. K. Neuhauser, R. Morgan, and S. L. Klein, "Impact of sex and gender on COVID-19 outcomes in Europe," *Biol Sex Differ*, vol. 11, no. 1, p. 29, Dec. 2020, doi: 10.1186/s13293-020-00304-9.

[27] M. O'Driscoll *et al.*, "Age-specific mortality and immunity patterns of SARS-CoV-2," *Nature*, vol. 590, no. 7844, pp. 140–145, Feb. 2021, doi: 10.1038/s41586-020-2918-0.

[28] A. B. Docherty *et al.*, "Features of 20 133 UK patients in hospital with covid-19 using the ISARIC WHO Clinical Characterisation Protocol: prospective observational cohort study," *BMJ*, p. m1985, May 2020, doi: 10.1136/bmj.m1985.

[29] C. E. Overton *et al.*, "Using statistics and mathematical modelling to understand infectious disease outbreaks: COVID-19 as an example," *Infect Dis Model*, vol. 5, pp. 409–441, 2020, doi: 10.1016/j.idm.2020.06.008.

[30] F. Zhou *et al.*, "Clinical course and risk factors for mortality of adult inpatients with COVID-19 in Wuhan, China: a retrospective cohort study," *The Lancet*, vol. 395, no. 10229, pp. 1054–1062, Mar. 2020, doi: 10.1016/S0140-6736(20)30566-3.

[31] L. Wynants *et al.*, "Prediction models for diagnosis and prognosis of covid-19: systematic review and critical appraisal," *BMJ*, p. m1328, Apr. 2020, doi: 10.1136/bmj.m1328.

[32] N. V. Chawla, K. W. Bowyer, L. O. Hall, and W. P. Kegelmeyer, "SMOTE: Synthetic Minority Over-sampling Technique," *Journal of Artificial Intelligence Research*, vol. 16, pp. 321–357, Jun. 2002, doi: 10.1613/jair.953.

[33] I. D. Apostolopoulos and T. A. Mpesiana, "Covid-19: automatic detection from X-ray images utilizing transfer learning with convolutional neural networks," *Phys Eng Sci Med*, vol. 43, no. 2, pp. 635–640, Jun. 2020, doi: 10.1007/s13246-020-00865-4.

[34] L. Yan *et al.*, "An interpretable mortality prediction model for COVID-19 patients," *Nat Mach Intell*, vol. 2, no. 5, pp. 283–288, May 2020, doi: 10.1038/s42256-020-0180-7.

[35] A. Vaid *et al.*, "Machine Learning to Predict Mortality and Critical Events in a Cohort of Patients With COVID-19 in New York City: Model Development and Validation," *J Med Internet Res*, vol. 22, no. 11, p. e24018, Nov. 2020, doi: 10.2196/24018.

[36] S. Saria *et al.*, "Development and Validation of ARC, a Model for Anticipating Acute Respiratory Failure in Coronavirus Disease 2019 Patients," *Crit Care Explor*, vol. 3, no. 6, p. e0441, Jun. 2021, doi: 10.1097/CCE.0000000000000441.